\pdfoutput=1
\documentclass[11pt]{article}

\usepackage{emnlp2021}
\usepackage{lipsum}
\newcommand\blfootnote[1]{%
  \begingroup
  \renewcommand\thefootnote{}\footnote{#1}%
  \addtocounter{footnote}{-1}%
  \endgroup
}

\usepackage{times}
\usepackage{latexsym}

\usepackage[T1]{fontenc}
\usepackage[utf8]{inputenc}

\usepackage{microtype}

\usepackage{array}
\usepackage{color}
\usepackage{bm}
\usepackage{enumitem}
\usepackage{booktabs}
\usepackage{amssymb}
\usepackage{amsfonts}
\usepackage{multirow}
\usepackage{graphicx}
\usepackage{amsmath}
\usepackage{CJKutf8}
\usepackage{url}
\usepackage[keeplayout]{covington}
\usepackage{bbding}
\usepackage{pifont}
\usepackage{wasysym}

\newcommand{\alns}[1] {\begin{align} #1 \end{align}}

\newcommand{\fscore}{\ensuremath{\mathrm{F}_{1}}}
\newcommand{\mask}{\texttt{[MASK]}}
\newcommand{\cls}{\texttt{[CLS]}}
\newcommand{\sep}{\texttt{[SEP]}}
\newcommand{\task}{\textsc{PZero}}
\newcommand{\prev}{\textsc{AS}}
\newcommand{\proposed}{\textsc{AS-PZero}}

\newcommand{\tokenemb}{\bm{e}^{\mathrm{token}}_{t}}
\newcommand{\posiemb}{\bm{e}^{\mathrm{position}}_{t}}
\newcommand{\addposiemb}{\bm{e}^{\mathrm{addposi}}_{t}}
\newcommand{\predemb}{\bm{e}^{\mathrm{predicate}}_{t}}

\title{Pseudo Zero Pronoun Resolution Improves\\ Zero Anaphora Resolution}

\author{
  \bf{Ryuto Konno}$^{\,1\ast}$ ~~
  \bf{Shun Kiyono}$^{\,2,3}$ ~~
    \bf{Yuichiroh Matsubayashi}$^{\,3,2}$ \\
  \bf{Hiroki Ouchi}$^{\,4,2}$ ~~
  \bf{Kentaro Inui}$^{\,3,2}$ \\
  ${}^{1}$Recruit Co., Ltd. ~ ${}^{2}$RIKEN ~ ${}^{3}$Tohoku University ~ ${}^{4}$Nara Institute of Science and Technology\\
  {\tt ryuto\_konno@r.recruit.co.jp}\\
  {\tt shun.kiyono@riken.jp} ~~ {\tt y.m@tohoku.ac.jp }\\
  {\tt hiroki.ouchi@is.naist.jp} ~~ {\tt inui@tohoku.ac.jp}\\
}

\begin{document}
\maketitle

%%%%%%%%%%
\begin{abstract}
Masked language models (MLMs) have contributed to drastic performance improvements with regard to zero anaphora resolution (ZAR).
To further improve this approach, in this study, we made two proposals.
The first is a new pretraining task that trains MLMs on anaphoric relations with explicit supervision, and the second proposal is a new finetuning method that remedies a notorious issue, the pretrain-finetune discrepancy.
Our experiments on Japanese ZAR demonstrated that our two proposals boost the state-of-the-art performance, and our detailed analysis provides new insights on the remaining challenges.
\end{abstract}

%%%%%%%%%%
%%%%%%%%%%
\section{Introduction}
\label{sec:introduction}
\begin{figure}[t]
    \centering
    \includegraphics[width=\hsize]{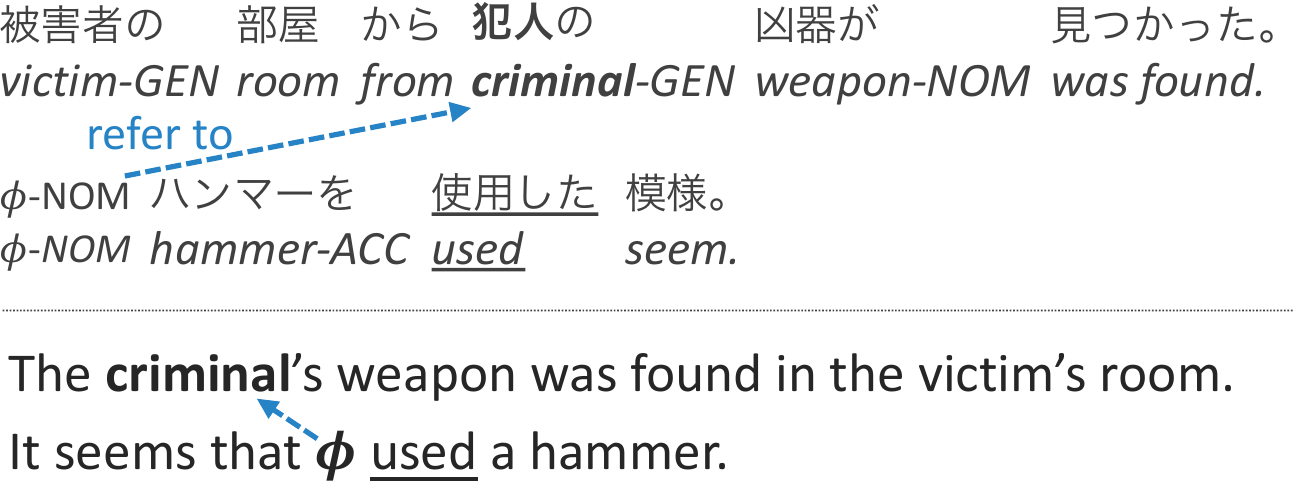}
    \caption{
        Example of argument omission in Japanese.
    }
    \label{fig:zar-example}
\end{figure}

\blfootnote{
    \hspace{-0.65cm}
    *Work done while at Tohoku University.
}
In pronoun-dropping languages such as Japanese and Chinese, the semantic arguments of predicates can be omitted from sentences.
As shown in Figure~\ref{fig:zar-example}, the semantic subject of the predicate \textit{used} is omitted and represented by $\phi$, which is called \textbf{zero pronoun}.
This pronoun refers to \textit{the criminal} in the first sentence.
This way, the task of recognizing the antecedents of zero pronouns is called \textbf{zero anaphora resolution} (ZAR).
This study focuses on Japanese ZAR. 

ZAR is a challenging task because it requires reasoning with commonsense knowledge about the semantic associations between zero pronouns and the local contexts of their preceding antecedents.
As shown in Figure~\ref{fig:zar-example}, to identify the omitted semantic subject of \textit{used}, the model should know the semantic relationship between \textit{the criminal's weapon} and {\it a hammer}, namely, \textit{a hammer is likely to be used as a weapon for murder and thus was used by the criminal}, is crucial.
We hereinafter refer to such knowledge as \textbf{anaphoric relational knowledge}.

A conventional approach to acquire anaphoric relational knowledge is to collect predicate--argument pairs from large-scale raw corpora and then, use them as features~\citep{sasano-etal-2008-fully,sasano2011discriminative,yamashiro-etal-2018-neural}, or using selectional preference probability \citep{shibata2018} in machine learning models.
A modern approach is to use masked language models (MLMs)~\citep{devlin:2019:NAACL}, which is effective in implicitly capturing anaphoric relational knowledge.
In fact, recent studies used pretrained MLMs and achieved drastic performance improvements in the tasks that require anaphoric relational knowledge, including Japanese ZAR~\cite{joshi-etal-2019-bert,aloraini-rlec-2020-zero,song-etal-2020-zpr2,konno-etal-2020-empirical}.

%%%%%
%%%%%
%%%%%
\begin{figure*}[t]
    \centering
    \includegraphics[width=\hsize]{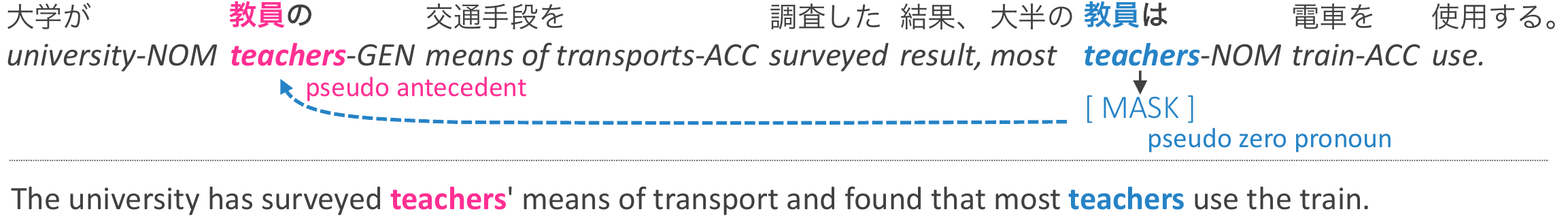}
    \caption{Example of our new pretraining task, \task{}. The second \textit{teachers} is regarded as a pseudo zero pronoun and is masked, and the first \textit{teachers} is its pseudo antecedent and should be selected to fill the mask.}
    \label{fig:selection-task-example}
\end{figure*}
%%%%%
To get more out of MLMs, in this paper, we propose a new training framework that pretrains and finetunes MLMs specialized for ZAR.
The idea is two-fold.

First, we design a new pretraining task that trains MLMs with explicit supervision on anaphoric relations.
Many current pretraining tasks adopt a form of the Cloze task, where each MLM is trained by predicting the original token filling the \mask{} token.
Although this task provides each MLM with no supervision on anaphoric relations, the MLM implicitly learns about them.
In contrast, our new task, called the \textbf{pseudo zero pronoun resolution} (\task{}), provides supervision on anaphoric relations.
\task{} assumes that when the same noun phrases (NPs) appear multiple times in a text, they are coreferent. 
From this assumption, we mask one of such multiple-occurring NPs as a \textbf{pseudo zero pronoun} and consider the other NPs as its \textbf{pseudo antecedents}.\footnote{In addition to antecedents, we deal with postcedents. We use the term ``antecedents" to refer both concepts for brevity.}
As shown in the example in Figure~\ref{fig:selection-task-example}, the NP, \textit{teachers}, appears twice.
The second is masked as a pseudo zero pronoun, and the first is regarded as its pseudo antecedent.
Then, given the masked zero pronoun, an MLM is trained to select its (pseudo) antecedent from the candidate tokens in the context.
The explicit supervision on such pseudo anaphoric relations allows MLMs to more effectively learn anaphoric relational knowledge.

Second, we address the issue called pretrain-finetune discrepancy~\cite{yang-neurips-2019-xlnet}.
Generally, some part of an MLM is changed for finetuning on a target task, e.g., discarding the pretrained parameters at the last layer or adding randomly-initialized new parameters.
Such changes in the architecture are known to be obstacles to the adaptation of pretrained MLMs to target tasks.
To solve this issue, we design a new ZAR model that takes over all the pretrained parameters of an MLM to the ZAR task with minimal modification.
This realizes a smoother adaptation of the anaphoric relational knowledge acquired during pretraining to ZAR.

Through experiments on Japanese ZAR, we verify the effectiveness of \task{} and the combination of \task{} and our new ZAR model.
Also, our analysis offers insights into the remaining challenges for Japanese ZAR.
To sum up, our main contributions are as follows:
\vspace{-0.2cm}
\begin{itemize}
\setlength{\parskip}{0cm} 
\setlength{\itemsep}{0cm}
\item We propose a new pretraining task, \task{}, that provides MLMs with explicit supervision on anaphoric relational knowledge;
\item We design a new ZAR model\footnote{Our code is publicly available: \url{ https://github.com/Ryuto10/pzero-improves-zar}} that makes full use of pretrained MLMs with minimal architectural modifications;
\item Our empirical results show that both the proposed methods can improve the ZAR performance and achieve state-of-the-art F$_1$ scores;
\item Our analysis reveals that the arguments far from predicates and the arguments of predicates in the passive voice are hard to predict.
\end{itemize}

%%%%%%%%%%
%%%%%%%%%%
\section{Japanese Zero Anaphora Resolution}
\label{sec:japanese-zar}

Japanese ZAR is often treated as a part of the predicate-argument structure analysis, which is the task of identifying semantic arguments for each predicate in a text.
In the NAIST Text Corpus (NTC)~\cite{iida2017naist}, a standard benchmark dataset that we used in our experiments, each predicate is annotated with the arguments filling either of the three most common argument roles: the nominative (\texttt{NOM}), accusative (\texttt{ACC}), or dative (\texttt{DAT}) roles.
If an argument of a predicate is a syntactic dependant of the predicate, we say that the argument is a syntactically dependent argument (DEP) and is relatively easy to identify.
If an argument of a predicate is omitted, in contrast, we say that the argument position is filled by zero pronouns.
This study is focused on recognizing such zero pronouns and identifying antecedents. 

The ZAR task is classified into the following three categories according to the positions of the arguments of a given predicate (i.e., the antecedent of a given zero pronoun):
\begin{description}
\setlength{\parskip}{0cm}
\setlength{\itemsep}{0cm}
    \item Intra-Sentential (\textit{intra}): the arguments \textbf{within the same sentence} where the predicate appears.
    \item Inter-Sentential (\textit{inter}): the arguments \textbf{in the sentences preceding} the predicate.
    \item Exophoric: the arguments (entities) that exist \textbf{outside the text}.
    These are categorized into one of three types: \textit{author}, \textit{reader}, and \textit{general}.\footnote{The definitions of \textit{Author} and \textit{reader} correspond to those in \citet{hangyo-etal-2013-japanese}. \textit{General} refers to the rest of \textit{exophoric}.} 
\end{description}
The identification of inter-sentential and exophoric arguments is an especially difficult task~\cite{shibata2018}.
For inter-sentential arguments, a model has to search the whole document.
For exophoric arguments, a model has to deal with entities outside the document.
Because of this difficulty, many previous studies have exclusively focused on the intra-sentential task.
In this paper, not only the intra-sentential task, we also treat inter-sentential and exophoric tasks as the same task formulations, as in previous studies.

%%%%%%%%%%
%%%%%%%%%%
\section{Pseudo Zero Pronoun Resolution}
\label{sec:pseudo-coreference-task}

%%%%%%%%%%
\subsection{Motivation and Task Formulation}
\label{subsec:pzero-motivation}
The proposed \task{} is a pretraining task for acquiring anaphoric relational knowledge necessary for solving ZAR.
\task{} is \textit{pseudo} since it is assumed that all the NPs with the same surface form have anaphoric relationships.
This assumption provides a large-scale dataset from raw corpora.
Although the assumption seems to be too strong, an empirical evaluation confirmed that the pretraining task was effective (Section~\ref{sec:results-analysis}).

The task is defined as follows: Let $\bm{X}$ be a given input token sequence $\bm{X} = (\bm{x}_1,\dots,\bm{x}_T)$ of length $T$, where one of the tokens is \mask{}.
Here, $\bm{x} \in \mathbb{R}^{\lvert \mathcal{V} \rvert}$ is a one-hot vector and $\mathcal{V}$ is a vocabulary set.
The task is to \textit{select} the token(s) corresponding to the original NP of \mask{} from the input tokens.
All the NPs with the same surface form as the masked NP are the answers of this task.

The most naive approach for masking NP is replacing all the tokens in the NP with the same number of \mask{} tokens.
However, this approach is not appropriate for acquiring anaphoric relational knowledge, as the model can simply use a superficial clue, that is, the number of \mask{} tokens, to predict the original NP.
Instead, we replace all the tokens in the NP with a \textbf{single} \mask{} token. 
Then, we formulate the task objective as predicting the last token in the original NP. 
This formulation is consistent with that of Japanese ZAR; when the argument consists of multiple tokens, the very last token is annotated as an actual argument.

%%%%%%%%%%
\subsection{Preparing Pseudo Data}
\label{subsec:data-gen}

To create training instances for \task{}, 
we first extract $n$ consecutive sentences from raw text and split them into a subword sequence.
We then insert \sep{} tokens as sentence separators~\cite{devlin:2019:NAACL}.
Subsequently, we prune tokens from the beginning of the sequence and then prepend \cls{} at the beginning.
As a result, the sequence consists of at most $T_{\mathrm{max}}$ subword tokens, which is the maximum input size of our model, as shown in Section~\ref{subsec:model-for-pretraining}.
Then, for each NP in the last sentence, we search for corresponding NPs with the same surface form in this sequence. 
Upon finding such NPs, we replace the selected NP in the last sentence with a single mask token and collect this sequence as a training instance.

%%%%%
\begin{figure*}[t]
    \centering
    \includegraphics[width=\hsize]{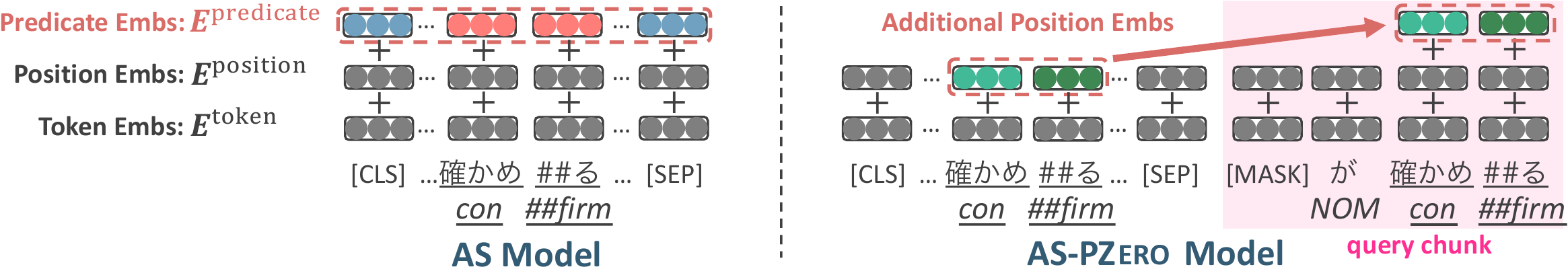}
    \caption{Input layer of \prev{} and \proposed{}. Their differences are that (1) a query chunk exists for \proposed{}, and (2) the position of the target predicate is informed via different embedding types: $\bm{E}^{\mathrm{predicate}}$ and $\bm{E}^{\mathrm{position}}$.}
    \label{fig:input_layers}
\end{figure*}
%%%%%

%%%%%%%%%%
\subsection{Pretraining Model}
\label{subsec:model-for-pretraining}
Our model for \task{} closely resembles that of the transformer-based MLM~\cite{devlin:2019:NAACL}.
Given an input sequence $\bm{X}$, each token $\bm{x}_t\in \{0,1\}^{|\mathcal{V}|}$ is mapped to an embedding vector of size $D$, namely, $\bm{e}_t\in \mathbb{R}^{D}$ as follows:
\alns{
    \bm{e}_{t} = \tokenemb{} + \posiemb{}. \label{equation:emb}
}
Here, an embedding vector $\tokenemb{} \in \mathbb{R}^{D}$ is obtained by computing $\tokenemb{} = \bm{E}^{\mathrm{token}} \bm{x}_t$, where $\bm{E}^{\mathrm{token}} \in \mathbb{R}^{D \times |\mathcal{V}|}$ is a token embedding matrix.
Similarly, an embedding vector $\posiemb{} \in \mathbb{R}^{D}$ is obtained from the position embedding matrix $\bm{E}^{\mathrm{position}} \in \mathbb{R}^{D \times T_{\mathrm{max}}}$ and a one-hot vector for position $t$.
$T_{\mathrm{max}}$ represents the predefined maximum input length of the model.

Then, the transformer layer encodes the input embeddings $\bm{e}_1,\dots,\bm{e}_{T}$ into the final hidden states $\bm{H}=(\bm{h}_1,\dots,\bm{h}_{T})$.
Given each hidden state $\bm{h}_t \in \mathbb{R}^{D}$ of the $t$-th token, we calculate the score $s_t \in \mathbb{R}$, which represents the likelihood that the token is a correct answer, by taking the dot product between the hidden state of the 
candidate token $\bm{h}_t$ and mask token $\bm{h}_{\rm mask}$:
\alns{
    s_t = (\bm{W}_{\rm 1} \bm{h}_t + \bm{b}_{\mathrm{1}})^{\top} \cdot (\bm{W}_{\rm 2} \bm{h}_{\rm mask} + \bm{b}_{\mathrm{2}}),
    \label{equation:pretraining-selection-score}
}
where $\bm{W}_{\rm 1}$ and $\bm{W}_{\rm 2}\in\mathbb{R}^{D \times D}$ are parameter matrices, and $\bm{b}_{\mathrm{1}}$, $\bm{b}_{\mathrm{2}} \in \mathbb{R}^D$ are bias terms.

We train the model to maximize the score of the correct tokens.
Specifically, we minimize the Kullback-Leibler (KL) divergence $\mathcal{L} = \mathrm{KL}(\bm{y}||\mathrm{softmax}(\bm{s}))$,
where $\bm{s} = (s_1, \dots, s_{T})$. $\bm{y} \in \mathbb{R}^{T}$ is the probability distribution of the positions of all the correct tokens. In this vector, the values corresponding to the positions of the correct tokens are set as $1/n$ and otherwise $0$, where $n$ is the number of correct tokens.

%%%%%%%%%%
%%%%%%%%%%
\section{ZAR Models}
\label{sec:model-for-zar}
Supposing that our model obtains anaphoric relational knowledge for ZAR by \textbf{pretraining} on \task{}, we design a ZAR model, as it can best utilize such knowledge during \textbf{finetuning}.
In this section, we first describe an argument selection model (Section~\ref{subsec:sequential-labeling-model}), which is considered the most straightforward adaptation of a pretrained model for ZAR.
Then, we propose a novel model that addresses the pretrain-finetune discrepancy (Section~\ref{subsec:selection-ft}).

%%%%%%%%%%
\subsection{Argument Selection with Label Probability: \prev{}}
\label{subsec:sequential-labeling-model}
The argument selection model, hereinafter \prev{}, is a model inspired by the model of \citet{kurita2018}.
From the recent standard practice of the pretrain-finetune paradigm~\cite{devlin:2019:NAACL}, we add a classification layer on top of the pretrained model.

The model takes an input sequence $\bm{X}$, which is created in a similar manner to that described in Section \ref{subsec:pzero-motivation}.
$\bm{X}$ consists of multiple sentences and is pruned to contain $T_{\mathrm{max}}$ tokens at maximum.
The target predicate is in the last sentence, and the \cls{} and \sep{} tokens are included.
Also, the model takes two natural numbers $p_{\mathrm{start}}$ and $p_{\mathrm{end}}$ as inputs, where $1 \leq p_{\mathrm{start}} \leq p_{\mathrm{end}} \leq T$.
These represent the position of the target predicate.

The model selects a filler token for each argument slot following a label assignment probability over $\bm{X}$: ${\rm argmax}_t P(t|\bm{X},l,p_{\mathrm{start}}, p_{\mathrm{end}})$, where $l \in \{{\tt NOM}, {\tt ACC}, {\tt DAT}\}$.
We regard \cls{} (i.e., $\bm{x}_{1}$) as a dummy token representing the case that the argument filler does not exist in the input sequence.
The model selects the dummy token in such cases.

The operation on the input layer of the model is shown on the left-hand side of Figure~\ref{fig:input_layers}.
First, each token $\bm{x}_t\in \{0,1\}^{|\mathcal{V}|}$ in a given input sequence ${\bm X}$ is mapped to an embedding vector $\bm{e}_t\in \mathbb{R}^{D}$ using the pretrained embedding matrices $\bm{E}^{\mathrm{token}}$ and $\bm{E}^{\mathrm{position}}$, and another new embedding matrix $\bm{E}^{\mathrm{predicate}} \in \mathbb{R}^{D\times 2}$, as follows:
\alns{
    \bm{e}_{t} &= \tokenemb{} + \posiemb{} +  \predemb{}, \label{equation:baseline-emb}
}
where $\bm{e}_t^{\mathrm{token}}$ and $\bm{e}_t^{\mathrm{position}}$ are the same as in Equation \ref{equation:emb}.
Moreover, $\bm{e}_t^{\mathrm{predicate}}$ is an embedding vector computed from $\bm{E}^{\mathrm{predicate}}$, $p_{\mathrm{start}}$, and $p_{\mathrm{end}}$.
This vector represents whether the token in position $t$ is a part of the predicate or not~\cite{he-etal-2017-deep}.

Second, we apply a pretrained transformer to encode each embedding $\bm{e}_t$ into the final hidden state $\bm{h}_t \in \mathbb{R}^{D}$.
The probability distribution of assigning the label $l$ over the input tokens $\bm{o}_{l}=(o_{l,1},...,o_{l,T}) \in \mathbb{R}^T$ is then obtained by the softmax layer:
\alns{
    o_{l,t} = \frac{
        \mathrm{exp}(\bm{w}_l^\mathsf{T}\bm{h}_t + b_l)
    }{
        \sum_{t}\mathrm{exp}(\bm{w}_l^\mathsf{T}\bm{h}_t + b_l)
    },
    \label{equation:baseline-sentence-wise}
}
where $\bm{w}_l \in \mathbb{R}^{D}$ and $b_l \in \mathbb{R}$.
Finally, from the probability distribution $\bm{o}_l$, the model selects the token with the maximum probability as the argument of the target predicate.

When the model selects the dummy token as an argument, we further classify the argument into the following four categories: $z \in \{\tt{author, reader, general, none}\}$.
Here, ${\tt none}$ shows no slot filler for this instance. 
The other three categories $\tt{author}$, ${\tt reader}$, and ${\tt general}$ represent that there is a certain filler entity but do not appear in the context (\textit{exophoric}).
For this purpose, we calculated a probability distribution over the four categories $\bm{o}^{\rm exo}_l=( o^{\rm exo}_{l,\tt{author}}, o^{\rm exo}_{l,\tt{reader}}, o^{\rm exo}_{l,\tt{general}}, o^{\rm exo}_{l,\tt{none}}) \in \mathbb{R}^4$ by applying a softmax layer to the hidden state of the dummy token $\bm{h}_1$ as follows:
\alns{
    o^{\rm exo}_{l, z} = \frac{
        \mathrm{exp}(\bm{w}_{l, z}^\mathsf{T}\bm{h}_1 + b_{l,z})
    }{
        \sum_{z}\mathrm{exp}(\bm{w}_{l, z}^\mathsf{T}\bm{h}_1 + b_{l,z})
    },
    \label{equation:baseline-exo}
}
where $\bm{w}_{l, z} \in \mathbb{R}^{D}$, and $b_{l,z} \in \mathbb{R}$.
Then the model selects the category with the maximum probability.

In the training step, we assign a gold label to the last token of an argument mention.
If there are multiple correct answers in the coreference relations in the context, we assign gold labels to all these mentions.
We prepare a probability distribution $\bm{y} \in \mathbb{R}^{T}$ of gold labels over the input token in a manner similar to that in Section~\ref{subsec:model-for-pretraining}.
The models are then trained to assign high probabilities to gold arguments.

%%%%%%%%%%
\subsection{Argument Selection as Pseudo Zero Pronoun Resolution: \proposed{}}
\label{subsec:selection-ft}
One potential disadvantage of the \prev{} model is that it may suffer from pretrain-finetune discrepancy. 
That is, \prev{} does not use the pretrained parameters, such as $\bm{W}_1$, $\bm{W}_2$, $\bm{b}_1$, and $\bm{b}_2$ in Equation~\ref{equation:pretraining-selection-score}, but is instead finetuned with randomly-initialized new parameters, such as $\bm{w}_l$ and $b_{l}$ in Equation~\ref{equation:baseline-sentence-wise}.
To make efficient use of the anaphoric relational knowledge acquired during pretraining, we resolve this discrepancy.
Inspired by studies addressing such discrepancies~\cite{Gururangan-acl-20,yang-neurips-2019-xlnet}, we propose a novel model for finetuning; argument selection as pseudo zero pronoun resolution (\proposed{}).

The underlying idea of \proposed{} is to solve ZAR as \task{}.
We use the network structure pretrained on \task{} as it is.
Thus, the parameters $\bm{w}_{l}$ and $b_{l}$ are no longer required.
To do this, we modify the input sequence $\bm{X}$ for ZAR and reformulate the ZAR task as \task{}.
Specifically, we prepare a short sentence, called a \textit{query chunk}, and append it to the end of the input sequence $\bm{X}$.
The query chunk represents a target predicate-argument slot whose filler is a single \mask{} token, so ZAR can be resolved by selecting the antecedent of the \mask{} token.

Let $\bm{X}^{\prime}$ denote the modified input of \proposed{}.
The input layer of the model is shown on the right-hand side of Figure~\ref{fig:input_layers}.
The query chunk consists of a \mask{} token, a token representing a target argument label (i.e., {\tt NOM}, {\tt ACC}, or {\tt DAT}), and a target predicate.
For example, when the number of tokens in the target predicate is represented as $T_{\mathrm{predicate}}=p_{\mathrm{end}}-p_{\mathrm{start}}+1$, the length of $\bm{X}^{\prime}$ is $T+2+T_{\mathrm{predicate}}$.
The modified input sequence is represented as $\bm{X}^{\prime}=(\bm{x}_1,\dots,\bm{x}_{T+2+T_{\mathrm{predicate}}})$.\footnote{The beginning of $\bm{X}^{\prime}$ is trimmed, so that the total number of tokens in $\bm{X}^{\prime}$ does not exceed the maximum input length of the model (i.e., $T_{\mathrm{max}}$) and $\bm{x}_1$ in $\bm{X}^{\prime}$ is \cls{}.}

Given a modified input sequence $\bm{X}^{\prime}$ and the start and end positions of the target predicate $p_{\mathrm{start}}, p_{\mathrm{end}} \in \mathbb{N}$, an input token $\bm{x}_t\in\{0,1\}^{|V|}$ is mapped to a token embedding $\bm{e}_t\in\mathbb{R}^D$ as follows:
\alns{
    \bm{e}_{t} &= \tokenemb{} + \posiemb{} +  \addposiemb{},
}
where $\addposiemb{}$ is an additional position embedding, which informs the model about the position of the target predicate.
This information is intended to be used for distinguishing the target predicate from the multiple predicates appearing with an identical surface form in the input sequence.
Specifically, for the target predicate in the query chunk, $\addposiemb{}$ is the same as the position embedding of the target predicate in the original sequence $\bm{X}$. Otherwise, $\addposiemb{}$ is zero:
\alns{
\addposiemb{}=
\begin{cases}
    \bm{e}^{\mathrm{position}}_{t^{\prime}} & (p_\mathrm{start}\leqq t^{\prime} \leqq p_\mathrm{end}) \\
    \bm{0} & (\mathrm{otherwise}),
\end{cases}
}
where $t^{\prime}=t-(T+3)+p_\mathrm{start}$.
For example, as shown in Figure~\ref{fig:input_layers}, the position embeddings of the target predicate (\textit{con} and \textit{\#\#firm}) are added to those in the query chunk.
Thus, we can avoid using the extra embedding matrix $\bm{E}^{\mathrm{predicate}}$ in Equation~\ref{equation:baseline-emb}.

We encode the embeddings with the transformer layer, and then use Equation~\ref{equation:pretraining-selection-score} for the remaining computation of \proposed{} to fill out the \mask{} token with the argument of the target predicate.
If the score of the dummy token ($\bm{x}_1$) is highest, the model computes \textit{exophoric} scores as described in Section~\ref{subsec:sequential-labeling-model} using Equation~\ref{equation:baseline-exo}.

%%%%%%%%%%
%%%%%%%%%%
\section{Experimental Settings}
\label{sec:experimental-settings}

\noindent\textbf{\task{} Dataset}\hspace*{3mm} 
Japanese Wikipedia corpus (Wikipedia) is the origin of the training data of \task{}.\footnote{We used the dump file as of September 1st, 2019 obtained from \href{https://dumps.wikimedia.org/jawiki/}{dumps.wikimedia.org/jawiki/}.}
All the NPs in the corpus are \task{} targets.
To detect NPs, we parsed Wikipedia using the Japanese dependency parser \textit{Cabocha}~\cite{cabocha} and applied a heuristic rule based on part-of-speech tags.
We used $n=4$ consecutive sentences to develop the input sequence $\bm{X}$.
From 17.4M sentences in Wikipedia, we obtained 17.3M instances as training data for \task{}.

\noindent\textbf{ZAR Dataset}\hspace*{3mm} 
For the ZAR task, we used NAIST Text Corpus 1.5 (NTC)~\cite{iida2010annotation,iida2017naist}, which is a standard benchmark dataset of this task~\cite{ouchi2017,matsubayashi:2018:COLING,omori2019,konno-etal-2020-empirical}.
We used the training, development, and test splits proposed by \citet{taira2008}.
The numbers of \textit{intra-sentential}, \textit{inter-sentential}, and \textit{exophoric} for the training/test instances were $18068/6159$, $11175/4081$, and $13676/3826$, respectively. 
The NTC details are shown in Appendix~\ref{sec:appendix-naist-text-corpus}.
The evaluation script corresponds to that of \citet{matsubayashi:2018:COLING}.

\noindent\textbf{Model}\hspace*{3mm} 
Our implementation is based on the Transformers library~\cite{Wolf2019HuggingFacesTS}. 
We used the pretrained parameters of the bert-base-japanese model as the initial parameters of our pretraining models.

We trained our model using an Adam optimizer~\cite{kingma:2015:ICLR} with warm-up steps.
As a loss function, we used cross-entropy for the Cloze task and prediction of \textit{exophoric} and used KL divergence for the rest. 
The details of the hyper-parameter search are in Appendix~\ref{sec:appendix-hyper-parameter-search}. 
Regarding the experiments on ZAR, we trained each model using five random seeds and reported the average score.

%%%%%
\begin{table}[t]
    \centering
    \small
    \tabcolsep 5pt
    \begin{tabular}{c|l|c|c|c}
    \toprule
       &         & ZAR            & DEP & All \\
    ID & Method  & \textit{intra} &     &     \\
    \midrule
    (a) & M\&I                       & 55.55 & 90.26 & 83.94 $\pm$ 0.12 \\
    (b) & O\&K                       & 53.50 & 90.15 & 83.82 $\pm$ 0.10 \\
    (c) & \citet{konno-etal-2020-empirical} & 64.15 & 92.46 & 86.98 $\pm$ 0.13 \\
    \midrule
    (d) & \prev{}                    & 69.32 & 93.65 & 88.87 $\pm$ 0.12 \\
    (e) & \proposed{}                & \textbf{69.91} & \textbf{93.83} & \textbf{89.06} $\pm$ 0.11 \\
    \midrule
    \end{tabular}
    \caption{\fscore{} scores on the NTC test set on \textbf{\textit{intra-sentential}} setting.
    M\&I: \citet{matsubayashi:2018:COLING}. O\&K: \citet{omori2019}.}
    \label{tab:intra-test-score}
\end{table}
%%%%%
\begin{table*}[t]
    \centering
    \small
    \begin{tabular}{c|c|cc|cc|cccc|c|c}
    \toprule
       & PT Task & \multicolumn{2}{c|}{Further PT Task} & \multicolumn{2}{c|}{FT Model} & \multicolumn{4}{c|}{ZAR}                                                            & DEP & All \\
    ID & Cloze &  Cloze & \task{}        & \prev{} & \proposed{}  & All & \textit{intra} & \textit{inter} & \textit{exophoric} &     &     \\
    \midrule
     (f) & \CheckmarkBold &   &   & \CheckmarkBold &   &          62.27 $\pm$ 0.42 &          71.55 &          44.30 &          64.04 &          94.44 &          82.97 \\
     (g) & \CheckmarkBold &   &   &   & \CheckmarkBold &          62.47 $\pm$ 0.53 &          71.09 &          45.20 &          64.41 &          94.46 &          83.03 \\
    \midrule    
     (h) & \CheckmarkBold & \CheckmarkBold &   & \CheckmarkBold &   &          62.54 $\pm$ 0.47 &          71.82 &          44.98 &          63.94 & \textbf{94.51} &          83.10 \\
     (i) & \CheckmarkBold & \CheckmarkBold &   &   & \CheckmarkBold &          62.85 $\pm$ 0.19 &          71.52 &          45.97 &          64.55 &          94.49 &          83.18 \\
     (j) & \CheckmarkBold &   & \CheckmarkBold & \CheckmarkBold &   &          63.06 $\pm$ 0.19 &          71.96 &          46.37 &          64.42 &          94.43 &          83.26 \\
     (k) & \CheckmarkBold &   & \CheckmarkBold &   & \CheckmarkBold & \textbf{64.18} $\pm$ 0.23 & \textbf{72.67} & \textbf{48.41} & \textbf{65.40} &          94.50 & \textbf{83.65} \\
    \bottomrule
    \end{tabular}
    \caption{\fscore scores in the NTC test set with the \textbf{\textit{inter-sentential}} setting.
    The bold values indicate the best results in the same column. PT and FT are abbreviations of pretraining and finetuning.
    The improvement of (k) over (h) is statistically significant in all the categories of ZAR \fscore{} ($p$ < 0.05) with a permutation test.}
    \label{tab:inter-test-score}
\end{table*}
%%%%%
\begin{table*}[t]
    \centering
    \tabcolsep 3.8pt
    \small
    \begin{tabular}{c|c|cc|cc|cc|c||cc|c}
    \toprule
                       & PT Task & \multicolumn{2}{c|}{Further PT Task} & \multicolumn{2}{c|}{FT Model} & \multicolumn{3}{c||}{\textit{intra}} & \multicolumn{3}{c}{\textit{inter}} \\
    \midrule
                    ID & Cloze &  Cloze & \task{} & \prev{} & \proposed{} &  Precision &   Recall &  \fscore{}  &  Precision &   Recall &  \fscore{} \\
    \midrule
    (h) & \CheckmarkBold & \CheckmarkBold &   & \CheckmarkBold &    & \textbf{76.52} &           67.67 &           71.82 $\pm$ 0.21 &           55.49 &           37.88 &           44.98 $\pm$ 1.05 \\
    (i) & \CheckmarkBold & \CheckmarkBold &   &   & \CheckmarkBold  &          75.59 &           67.87 &           71.52 $\pm$ 0.22 &           56.31 &           38.91 &           45.97 $\pm$ 0.42 \\
    (j) & \CheckmarkBold &   & \CheckmarkBold & \CheckmarkBold &    &          75.85 &           68.46 &           71.96 $\pm$ 0.38 &           55.92 &           39.61 &           46.37 $\pm$ 0.34 \\
    (k) & \CheckmarkBold &   & \CheckmarkBold &   & \CheckmarkBold  &          76.06 &  \textbf{69.58} &  \textbf{72.67} $\pm$ 0.32 &  \textbf{57.63} &  \textbf{41.74} &  \textbf{48.41} $\pm$ 0.35 \\
    \bottomrule
    \end{tabular}
        \caption{The NTC test set results of the \textbf{\textit{inter-sentential}} setting. The bold values indicate the best results in the same column group. PT and FT are abbreviations of pretraining and finetuning.}
        \label{tab:test-scores-details}
\end{table*}
\begin{table*}[t]
    \centering
    \small
    \tabcolsep 4.5pt
        \begin{tabular}{c|l|rr|r||rr|r}
        \toprule
        {} & {} & \multicolumn{2}{c|}{\textit{intra} recall} & \# of  & \multicolumn{2}{c|}{\textit{inter} recall} & \# of  \\
        ID & Type of instances & Model (h) & Model (k) & {instances} & Model (h) & Model (k) & {instances} \\
        \midrule
        \multicolumn{8}{c}{\textbf{(I) Number of gold antecedents in input}} \\
        \midrule
        (1) & Only one            &  65.87 &  69.12 &  2001 &  35.96 &  39.57 &  1218 \\
        (2) & More than one       &  73.78 &  74.26 &  1247 &  53.1 &  54.1 &   872 \\
        \midrule
        \multicolumn{8}{c}{\textbf{(II) Position of the argument relative to the target predicate}} \\
        \midrule
        (3) & One sentence before            &      - &      - &     0 &  48.86 &  51.96 &  1099 \\
        (4) & Two sentences before           &      - &      - &     0 &  37.7 &  42.6 &   411 \\
        (5) & More than two sentences before &      - &      - &     0 &  40.5 &  40.3 &   516 \\
        (6) & Out of input sequence         &      - &      - &     0 &  0.0 &  0.0 &    64 \\
        \midrule
        \multicolumn{8}{c}{\textbf{(III) Voice of the target predicate}} \\
        \midrule
        (7) & Active                           &  70.80 &  72.62 &  2918 &  45.02 &  47.42 &  1877 \\
        (8) & Passive                          &  51.8 &  57.9 &   309 &  25.2 &  29.1 &   206 \\
        (9) & Causative                        &  55 &  50 &    20 &  60 &  60 &     7 \\
        (10) & Causative \& Passive               &    100 &    100 &     1 &      - &      - &     0 \\
        \midrule
            & All                                           &   68.9 &  \textbf{71.09} &  3248 &  43.11 &  \textbf{45.65} &  2090 \\
        \bottomrule
        \end{tabular}
    \caption{Recall scores for each type of instance in the NTC development set. Models (h) and (k) are from Table~\ref{tab:inter-test-score}.}
    \label{tab:case-study}
\end{table*}
%%%%%

%%%%%%%%%%
%%%%%%%%%%
\section{Results and Analysis}
\label{sec:results-analysis}
We have two distinct goals in this experiment, that is, to investigate the effectiveness of (1) pretraining on \task{} and (2) finetuning on \proposed{}.
To achieve these goals, we first compare our \prev{} and \proposed{} models with previous studies to ensure that our models are strong enough in a conventional experimental setting, i.e., the \textit{intra-sentential} setting (Section~\ref{subsec:intra-sentential}).
Then we investigate (1) and (2) based on \textit{inter-sentential} setting (Section~\ref{subsec:inter-sentential}).

%%%%%%%%%%
\subsection{Intra-sentential Experiment}
\label{subsec:intra-sentential}
In this setting, the input sequence consists of a \textbf{single sentence}, and
only the \textit{intra}-zero and DEP arguments are targets of the evaluation.
As mentioned in Section~\ref{sec:japanese-zar}, most of the previous studies on Japanese ZAR use this setting~\cite{matsubayashi:2018:COLING,omori2019,konno-etal-2020-empirical}.
Thus, we can strictly compare our results with those of other studies in this setting.

We finetuned \prev{} and \proposed{} from a pretrained MLM.
The results in Table~\ref{tab:intra-test-score} show that both the \prev{} and \proposed{} models already outperformed the previous state-of-the-art models in \textit{intra}-zero and DEP~\cite{konno-etal-2020-empirical} with large margins.
This improvement is due to the difference in the use of the pretrained MLM; given a pretrained MLM, we finetuned its entire parameters whereas \citet{konno-etal-2020-empirical} used it as input features.
Additionally, our pretrained MLM was trained better than theirs.

%%%%%%%%%%
\subsection{Inter-sentential Experiment}
\label{subsec:inter-sentential}
In this setting, the input sequence consists of \textbf{multiple sentences}: a sentence containing a target predicate and preceding sentences in the document.
The \textit{intra-sentential}, \textit{inter-sentential}, \textit{exophoric}, and DEP arguments are the evaluation targets.

We investigate the effectiveness of the proposed \task{} and \proposed{}.
For the experiment, we initialized the parameters of the transformer-based model with the pretrained MLM (pretrain 1) and further pretrained the model on Cloze and \task{} with the same number of updates.
This resulted in having two pretrained models (pretrain 2 \& 3).
Then, we created models of all the possible combinations from \{pretrain 1, 2, 3\} and \{\prev{}, \proposed{}\}, resulting in the six models shown in Table~\ref{tab:inter-test-score}.

\paragraph{(I) Do inter-sentential contexts help intra-sentential argument identification?}
We first investigate the impact of inter-sentential context on the performances of \textit{intra}-zero and DEP by comparing the models (f) and (g) in Table~\ref{tab:inter-test-score} and the models (d) and (e) in Table~\ref{tab:intra-test-score}.
Here, note that model architectures of (f) and (g) are identical to those of (d) and (e), respectively.
In addition, the evaluation instances of the \textit{intra}-zero and DEP categories are the same for all four models.
The differences are that the models (f) and (g) have broader contexts (inter-sentential contexts), i.e., multiple preceding sentences as inputs, and extra training signals from both the \textit{inter}-zero and \textit{exophoric} instances.
A comparison of these four models shows that (f) and (g) have better performance than (d) and (e) in \textit{intra}-zero and DEP.
This result indicates that inter-sentential contexts are important clues even for identifying intra-sentential argument relations.
This result is consistent with those of \citet{guan-etal-2019-semantic} and \citet{shibata2018}, which discussed a method for utilizing inter-sentential contexts as clues for resolving semantic relations in target sentences.

\paragraph{(II) Does pretraining on PZero improve the performance of AS?}
As shown in Table~\ref{tab:inter-test-score}, the comparison between the models pretrained on \task{} (j) and Cloze (h) shows that \task{} outperforms Cloze, especially in \textit{inter}-zero argument (44.98 $\rightarrow$ 46.37).
As discussed in Section~\ref{sec:japanese-zar}, \textit{inter}-zero is challenging because there are multiple answer candidates across the sentences.
The improvement in \textit{inter}-zero implies that the model effectively learns anaphoric relational knowledge through the pretraining on \task{}.

\paragraph{(III) Does pretraining on PZero improve the performance of AS-PZero?}
The performance comparison between the models (j) and (k) demonstrates the effectiveness of the combination of \task{} and \proposed{}.
The model (k) achieved the best result in all categories except for DEP.
This indicates that \proposed{} has successfully addressed the pretrain-finetune discrepancy and that it effecively used the anaphoric relational knowledge learned from \task{}.

Table~\ref{tab:test-scores-details} shows the precision and recall of the models (h)-(k) for the \textit{intra}-zero and \textit{inter}-zero arguments.
The model (k) achieved the best recall performance in both categories and indicated that the proposed \task{} contributes mainly to the improvement in recall.

%%%%%%%%%%
\subsection{Analysis}
\label{subsec:analysis}
We analyze the source of the improvement in recall as observed in Table~\ref{tab:test-scores-details}.
Table~\ref{tab:case-study} shows our analysis of the \textit{intra}/\textit{inter-sentential} arguments from three aspects \textbf{I--III} and compares the detailed results of the baseline model (h) and our model (k).

\noindent\textbf{(I) Number of gold antecedents in input}\hspace*{3mm}
This number determines the difficulty of the anaphora resolution.
This is because the saliency of the entity is an important clue, i.e., ZAR is difficult when the argument appears only once in an input.
Our model improved the performance of such difficult instances by a large margin (65.87 $\rightarrow$ 69.12 in {\it intra}-zero and 35.96 $\rightarrow$ 39.57 in {\it inter}-zero).

\noindent\textbf{(II) Position of the argument relative to the target predicate}\hspace*{3mm} 
The distance between a predicate and its argument determines the difficulty of ZAR.
According to (3), (4), and (5), the performance of \textit{inter-sentential} decreased as the predicate was farther from its argument's last surface-form appearance.
Interestingly, the performance of the two models was comparable in (5), which is the case that the arguments are more than two sentences away from the target predicate.
This result indicates that the proposed method is not effective for these instances.
The error analysis on these instances revealed the fact that even though the argument did not appear explicitly, it was semantically present throughout the context as omitted arguments of the multiple predicates, all pointing to the same entity.
This suggests that combining our proposed model with a model that propagates ZAR results through relevant contexts~\cite{shibata2018} can further improve ZAR performance.

\noindent\textbf{(III) Voice of the target predicate}\hspace*{3mm}
Identifying the arguments of the predicate in the nonactive voice clause is difficult because of {\it case alternation}; semantic subjects and objects appear in the other syntactic positions.
Table~\ref{tab:case-study} shows that both models perform worse in (8) and (9) than in (7).
Also the case alternation is different for every predicate.
Thus, the model had to learn each behavior from training data and raw corpus.
However, acquiring such information is not in the scope of \task{}.

%%%%%%%%%%
%%%%%%%%%%
\subsection{Discussion on Pseudo Data Generation}
\label{subsec:discussion}
In this study, we generated pseudo data for \task{} by exploiting the strong assumption that all the NPs with the same surface form have anaphoric relationships (Section~\ref{subsec:pzero-motivation}).
The advantage of our method is its high scalability in data collection; we can obtain a large amount of pseudo instances from raw corpora.
Our empirical evaluation showed that our assumption is effective, however more sophisticated methods could be considered.
Our future work includes analyzing the noise in pseudo data, i.e., NPs with the same surface but \emph{no} anaphoric relationships, and its effect on the model performance.

%%%%%%%%%%
%%%%%%%%%%
\section{Related Work}
\label{sec:related-work}

\noindent\textbf{Anaphoric Relational Knowledge}\hspace*{3mm}
Our proposed pretraining task for acquiring anaphoric relational knowledge is related to script knowledge acquisition~\cite{chambers:2009}.
Script knowledge models chains of typical events (predicates and their arguments).
Between events, some arguments are shared and represented as variables, such as \texttt{purchase} X $\rightarrow$ \texttt{acquire} X, which can be regarded as a type of anaphoric relational knowledge.
While script knowledge only deals with shared arguments as anaphoric (coreferring) phenomena, anaphoric relational knowledge is not limited to them.
In the sentence of Figure~\ref{fig:zar-example}, the word \textit{criminal} is not an argument of the predicate and is ignored in script knowledge, whereas it is within the scope of this work.
Thus, it can be said that this work deals with broader anaphoric phenomena. 

\noindent\textbf{Zero Anaphora Resolution (ZAR)}\hspace*{3mm} 
ZAR has been studied in multiple languages, such as Chinese~\cite{yin2018zero}, Japanese~\cite{iida:2016:intra}, Korean~\cite{han2006korean}, Italian~\cite{iida-poesio-2011-cross}, and Spanish~\cite{palomar2001algorithm}.
ZAR faces a lack of labeled data, which is a major challenge, and the traditional approach to overcome this is to use large-scale raw corpora.
Several studies have employed these corpora as a source of knowledge for ZAR, e.g., case-frame construction~\cite{sasano-etal-2008-fully,sasano2011discriminative,yamashiro-etal-2018-neural} and selectional preference probability~\cite{shibata2016}.
Furthermore, semi-supervised learning approaches, such as pseudo data generation~\cite{liu:2017} and adversarial training~\cite{kurita2018}, have been proposed.
However, the use of pretrained MLM has been the most successful approach~\cite{konno-etal-2020-empirical}, and we sought to improve the pretraining task to better acquire anaphoric relational knowledge.

\noindent\textbf{Pseudo Zero Pronoun Resolution (\task{})}\hspace*{3mm}
Several studies have created training instances in a similar way as in \task{}.
For example, \citet{liu:2017} casted the ZAR problem as a reading comprehension problem, such that the model chose an appropriate word for the \mask{} from the vocabulary set.
The difference is that, unlike their work, we filled the \mask{} by selecting a token from the given sentences.
Also, \citet{Kocijan-acl-19} created similar training data for Winograd Schema Challenge~\cite{levesque-aaai-11-winograd}.
While we considered replacing arbitrary NPs with \mask{}, they exclusively replaced the personal name.
We expect that our approach is more suitable for ZAR because arguments are not necessarily personal names.

\noindent\textbf{Pretrain-finetune Discrepancy}\hspace*{3mm}
Addressing the discrepancy between pretraining and finetuning is one of the successful approaches for improving the use of pretrained MLMs.
For example, \citet{Gururangan-acl-20} addressed the discrepancy with respect to the domain of the training dataset.
Furthermore, \citet{yang-neurips-2019-xlnet} indicated that \mask{} is used during the pretraining of MLM but never during finetuning.
They improved a model architecture to mitigate such discrepancies.
Therefore, inspired by these studies, we designed a finetuning model (\proposed{}) that is suitable for a model pretrained on \task{} and demonstrated its effectiveness.

\noindent\textbf{Prompt-based Learning}\hspace*{3mm}
Our use of query chunk in \proposed{} can be seen as a prompt-based learning approach~\cite{radford-etal-2019-language,brown-etal-2020-GPT3}, which has been actively studied~\citep{liu2021pre}.
In a typical prompt-based learning with a pretrained MLM, a model is trained to replace the masked token with a token from a predefined vocabulary~\cite{shick-etal-2020-automatically,shich-schutze-2021-exploitCloze,shick-schutze-2021-itsNotJust,gao-etal-2021-makingPtrLM}.
Our model is pretrained on \task{}, which is a task to select a pseudo antecedent from the preceding context.
Thus, we designed \proposed{} as a model to select the argument from the input sentences using a prompt-based approach to avoid the pretrain-finetune discrepancy.

%%%%%%%%%%
%%%%%%%%%%
\section{Conclusion}
\label{sec:conclusion}
In this study, we proposed a new pretraining task, \task{}, which aims to explicitly teach the model anaphoric relational knowledge necessary for ZAR.
We also proposed a ZAR model to remedy the pretrain-finetune discrepancy. 
Both the proposed methods improved the performance of Japanese ZAR, leading to a new state-of-the-art performance.
Our analysis suggests that the hard subcategories of ZAR; distant arguments and passive predicates are still challenging.

%%%%%%%%%%
%%%%%%%%%%
\section*{Acknowledgements}
We thank anonymous reviewers for their insightful comments.
We thank Jun Suzuki, Ana Brassard, Tatsuki Kuribayashi, Takumi Ito, Shiki Sato, and Yosuke Kishinami for their valuable comments.
This work was supported by JSPS KAKENHI Grant Numbers JP19K12112, JP19K20351, and JP19H04162.

\bibliography{anthology,custom}
\bibliographystyle{acl_natbib}

\clearpage
\appendix

\begin{table}[t]
    \centering
    \tabcolsep 3pt
    \small
        \begin{tabular}{ll|rrrr}
        \toprule
        Dataset &    {} & \textit{dep} & \textit{intra} & \textit{inter} & \textit{exophoric} \\
        \midrule
          Training &   NOM &  36934 &  12219 &   7843 &  11511 \\
                    &   ACC &  24654 &   2136 &    948 &    128 \\
                    &   DAT &   5744 &    465 &    294 &     60 \\
        \midrule
        Development &   NOM &   7424 &   2665 &   1812 &   1917 \\
                    &   ACC &   5055 &    445 &    177 &     32 \\
                    &   DAT &   1612 &    138 &    101 &     28 \\
        \midrule
              Test &   NOM &  14003 &   4993 &   3565 &   3717 \\
                    &   ACC &   9407 &    906 &    371 &     55 \\
                    &   DAT &   2493 &    260 &    145 &     54 \\
        \bottomrule
        \end{tabular}
    \caption{Statistics of NAIST Text Corpus 1.5}
    \label{tab:appendix-ntc-stats}
\end{table}
%%%%%
\begin{table*}[t]
    \centering
    \small
    \tabcolsep 4pt
        \begin{tabular}{l|l}
        \toprule
        \textbf{Configurations} & \textbf{Values} \\
        \midrule
        Optimizer               & Adam~\cite{kingma:2015:ICLR} ($\beta_1=0.9$,$\beta_2=0.999$,$\epsilon=1\times10^{-8}$) \\
        Hidden State Size ($D$) & 768 (defined in \texttt{bert-base-japanese}) \\
        $T_{\mathrm{max}}$      & 512 (defined in \texttt{bert-base-japanese}) \\
        \midrule
        \midrule
        \multicolumn{2}{c}{\textbf{Further Pretraining}} \\
        \midrule
        Mini-batch Size         & 2,048 \\
        Max Learning Rate       & $1.0 \times 10^{-4}$ (Cloze task), $2.0 \times 10^{-5}$ (\task{}) \\
        Learning Rate Schedule  & Inverse square root decay \\
        Warmup Steps            & 5,000 \\
        Number of Updates       & 30,000 \\
        Loss Function           & Cross entropy (Cloze task) and KL divergence (\task{}) \\
        MLM's Mask Position     & Random for each epoch \\
        \midrule
        \midrule
        \multicolumn{2}{c}{\textbf{Finetuning}} \\
        \midrule
        Mini-batch Size         & 256 \\
        Max Learning Rate       & $1.0 \times 10^{-4}$ (\prev{}) and $1.0 \times 10^{-5}$ (\proposed{}) \\
        Learning Rate Schedule  & Same as described in Appendix A of \citet{matsubayashi:2018:COLING} \\
        Number of Epochs        & 150   \\
        Stopping Criterion      & Same as described in Appendix A of \citet{matsubayashi:2018:COLING} \\
        Loss Function           & KL divergence (\prev{}) and KL divergence (\proposed{}) and Cross entropy (prediction of \textit{exophoric}) \\
        \bottomrule
        \end{tabular}
    \caption{List of hyper-parameters}
    \label{tab:appendix-hypara-all}
\end{table*}
%%%%%
\begin{table*}[t]
    \centering
    \footnotesize
        \begin{tabular}{ll|l}
        \toprule
	    \textbf{Pretraining Task} & \textbf{Finetuning Model} & \textbf{Loss Function}                            \\
	    \midrule
	    \textit{Cloze}    & -                      &	\{\textbf{Softmax Cross Entropy}\}	              \\
	    \task{}           & -                &	\{Sigmoid Cross Entropy, \textbf{KL Divergence}\} \\
	    -       & \prev{}      &  \{Sigmoid Cross Entropy, \textbf{KL Divergence}\} \\ 
	    -       & \proposed{}                &	\{Sigmoid Cross Entropy, \textbf{KL Divergence}\} \\
        \midrule
        \midrule
	     & & \textbf{Maximum Learning Rate}   \\
	    \midrule
	    \textit{Cloze}    & -  &	\{$\bm{1.0\times10^{-4}}$, $5.0\times10^{-5}$, $2.0\times10^{-5}$, $1.0\times10^{-5}$, $5.0\times10^{-6}$\} \\
	    \task{}           & -  &	\{$1.0\times10^{-4}$, $5.0\times10^{-5}$, $\bm{2.0\times10^{-5}}$, $1.0\times10^{-5}$, $5.0\times10^{-6}$\} \\
	    - & \prev{} & \{$\bm{1.0\times10^{-4}}$, $5.0\times10^{-5}$, $2.0\times10^{-5}$, $1.0\times10^{-5}$\} \\
	    - & \proposed{} &	\{$1.0\times10^{-4}$, $\bm{5.0\times10^{-5}}$, $2.0\times10^{-5}$, $1.0\times10^{-5}$\} \\
        \bottomrule
        \end{tabular}
    \caption{Candidates of hyper-parameters. \textbf{Bold} value indicates the adopted values.}
    \label{tab:appendix-hypara-search-params}
\end{table*}
%%%%%
\begin{table*}[t]
    \centering
    \footnotesize
        \begin{tabular}{l|r||cccc|c|c}
        \toprule
         & & \multicolumn{4}{c|}{ZAR} & DEP & All \\
        Loss Function & Learning Rate & All & \textit{intra} & \textit{inter} & \textit{exophoric} & & \\
        \midrule
        \midrule
        \multicolumn{8}{c}{Argument Selection as \task{} (\proposed{})}\\
        \midrule
        Sigmoid Cross Entropy & $1.0\times10^{-4}$ &   61.66 &    71.10 &    45.29 &  61.98 &  94.64 &  83.07 \\
        & $5.0\times10^{-5}$ &   62.00 &    71.52 &    45.25 &  62.80 &  94.80 &  83.25 \\
        & $2.0\times10^{-5}$ &   62.28 &    71.26 &    47.25 &  62.15 &  94.47 &  83.23 \\
        & $1.0\times10^{-5}$ &   62.20 &    71.17 &    48.02 &  61.57 &  94.54 &  83.22 \\
        \midrule
        \textbf{KL Divergence} & $1.0\times10^{-4}$ &   63.30 &    72.40 &    47.33 &  64.21 &  94.82 &  83.74 \\
        & $\bm{5.0\times10^{-5}}$ &   \textbf{63.73} &    \textbf{72.55} &    \textbf{48.39} &  \textbf{64.33} &  \textbf{94.99} &  \textbf{83.97} \\
        & $2.0\times10^{-5}$ &   62.52 &    71.78 &    46.98 &  62.96 &  94.91 &  83.52 \\
        & $1.0\times10^{-5}$ &   61.60 &    70.38 &    47.29 &  61.44 &  94.45 &  82.86 \\
        \midrule
        \midrule
        \multicolumn{8}{c}{Argument Selection with Label Probability (\prev{})} \\
        \midrule
        \textbf{KL Divergence} & $\bm{1.0\times10^{-4}}$ &   \textbf{62.77} & \textbf{72.68} & 45.65 & \textbf{63.15} & \textbf{95.03} &  \textbf{83.70} \\
        & $5.0\times10^{-5}$ &   62.28 &    71.87 &    \textbf{46.65} &  62.39 &  94.96 &  83.48 \\
        & $2.0\times10^{-5}$ &   61.73 &    71.58 &    45.99 &  61.56 &  94.52 &  82.97 \\
        & $1.0\times10^{-5}$ &   61.57 &    71.60 &    45.57 &  61.22 &  94.73 &  83.04 \\
        \midrule
        Sigmoid Cross Entropy& $1.0\times10^{-4}$ &   59.25 &    69.02 &    41.30 &  60.42 &  94.36 &  82.05 \\
        & $5.0\times10^{-5}$ &   57.51 &    66.88 &    40.34 &  58.47 &  93.90 &  81.20 \\
        & $2.0\times10^{-5}$ &   59.29 &    68.87 &    42.33 &  60.16 &  94.24 &  82.07 \\
        & $1.0\times10^{-5}$ &   59.40 &    68.90 &    43.28 &  59.77 &  94.47 &  82.22 \\
        \bottomrule
        \end{tabular}
    \caption{
        \fscore{} scores on the NTC \textbf{development set} for hyper-parameter search. Bold value indicates the best results in the same column.
    }
    \label{tab:appendix-hypara-search-on-dev}
\end{table*}
%%%%%

%%%%%%%%%%
%%%%%%%%%%
\section{Statistics of NAIST Text Corpus 1.5}
\label{sec:appendix-naist-text-corpus}
We used NAIST Text Corpus 1.5 (NTC)~\cite{iida2010annotation,iida2017naist} for ZAR task.
Table~\ref{tab:appendix-ntc-stats} shows the number of instances in NTC.

\section{Hyperparameter Search on Validation Set}
\label{sec:appendix-hyper-parameter-search}
Table~\ref{tab:appendix-hypara-all} shows a complete list of hyper-parameters used in this study.
For both pretraining and finetuning, maximum learning rate and loss function are the target of the hyperparameter search.
All the candidates of learning rates and loss functions are presented in Table~\ref{tab:appendix-hypara-search-params}. 
We used Nvidia Tesla V100 for the entire experiment.

\noindent\textbf{Pretraining on Cloze}\hspace*{3mm} 
For the hyperparameter search of \textit{Cloze} task, we adopted the hyperparameters that achieves the lowest perplexity value.
We adopted $1.0\times 10^{-4}$ for maximum learning rate.
We used the development set that we created from Japanese Wikipedia.

\noindent\textbf{Pretraining on \task{}}\hspace*{3mm} 
For the hyperparameter search of \task{}; the parameters were determined by the validation performance on \task{} and ZAR.\footnote{The task formulation of \task{} and \proposed{} are quite similar. Thus we can evaluate the model, which is pretrained on \task{}, directly on ZAR without finetuning.}
We eventually employed the parameters with the highest \fscore{} on \textit{inter} arguments.

\noindent\textbf{Finetuning on ZAR}\hspace*{3mm} 
For the hyperparameter search of ZAR; the hyperparameters that achieve the highest overall ZAR \fscore{} were used.
Here, we finetuned pretrained MLM without any further pretraining.
Table~\ref{tab:appendix-hypara-search-on-dev} shows the result of our search process.
%%%%%
\begin{table}[t]
    \centering
        \footnotesize
        \begin{tabular}{p{32mm}|rr}
        \toprule
                  &   Training  & Development \\
        \midrule
        Documents &   1,121,217 &   300 \\
        Sentences &  17,436,975 & 3,622 \\
        Instances of \task{} &  17,353,590 & 3,236 \\
        \bottomrule
        \end{tabular}
    \caption{Statistics of Japanese Wikipedia}
    \label{tab:appendix-raw-stats}
\end{table}
%%%%%

%%%%%%%%%%
%%%%%%%%%%
\section{Heuristics for Extracting Noun Phrases from Raw Text}
\label{sec:appendix-noun=phrase-extraction}
In order to extract noun phrases (NPs) from Japanese Wikipedia, we first parsed the corpus using Japanese dependency parser \textit{Cabocha}~\cite{cabocha}.
The parser divides the sentences into a phrase (Japanese ``bunsetsu").
Note that each bunsetsu consists of a sequence of words.
We then extracted the NPs as follows:

\begin{enumerate}
    \setlength{\parskip}{0cm}
    \setlength{\itemsep}{0cm}
  \item Choose a phrase that (1) contains noun(s) and (2) does not contain verb(s).
  \item Scan the phrase from the end, and keep eliminating words until a noun appears.
  \item Scan the phrase from the beginning, and keep eliminating words until a word other than a symbol appears.
  \item The remaining words are regarded as a noun phrase. If the remaining words contain symbols, alphabet, or numbers only, then the words are not discarded.
\end{enumerate}

Table~\ref{tab:appendix-raw-stats} shows the statistics of Japanese Wikipedia and the number of \task{} instances generated from this process.

%%%%%%%%%%
%%%%%%%%%%
\section{Performance on Validation Set}
We report the performance on development set of NTC for model (d) to model (k), in Table~\ref{tab:intra-dev-score} and Table~\ref{tab:inter-dev-score}.
Here, each model ID follows that of Table~\ref{tab:intra-test-score} and Table~\ref{tab:inter-test-score}.
%%%%%
\begin{table}[t]
    \centering
    \small
    \tabcolsep 5pt
    \begin{tabular}{c|l|c|c|c}
    \toprule
       &         & ZAR            & DEP & All \\
    ID & Method  & \textit{intra} &     & \\
    \midrule
    (d) & \prev{}                    & 70.41 &  94.21 &  89.48$\pm$0.10 \\
    (e) & \proposed{}                & \textbf{70.66} &  \textbf{94.23} &  \textbf{89.50}$\pm$0.08 \\
    \midrule
    \end{tabular}
    \caption{
        \fscore{} scores on the NTC development set on \textbf{\textit{intra-sentential}} setting.
    }
    \label{tab:intra-dev-score}
\end{table}
%%%%%
\begin{table*}[t]
    \centering
    \small
    \begin{tabular}{c|c|cc|cc|cccc|c|c}
    \toprule
       & PT Task & \multicolumn{2}{c|}{Further PT Task} & \multicolumn{2}{c|}{FT Model} & \multicolumn{4}{c|}{ZAR}                                                            & DEP & All \\
    ID & Cloze &  Cloze & \task{}        & \prev{} & \task{}  & All & \textit{intra} & \textit{inter} & \textit{exophoric} &     &     \\
    \midrule
     (f) & \CheckmarkBold &   &   & \CheckmarkBold &   &          62.97$\pm$0.27 &    72.67 &    46.84 &  63.33 &  94.99 &  83.73 \\
     (g) & \CheckmarkBold &   &   &   & \CheckmarkBold &          63.34$\pm$0.16 &    72.61 &    47.21 &  64.04 &  94.96 &  83.83 \\
    \midrule    
     (h) & \CheckmarkBold & \CheckmarkBold &   & \CheckmarkBold &   &          63.42$\pm$0.29 &    72.85 &    48.26 &  63.21 &  \textbf{95.02} &  83.92 \\
     (i) & \CheckmarkBold & \CheckmarkBold &   &   & \CheckmarkBold &          63.63$\pm$0.29 &    72.57 &    48.70 &  63.82 &  94.95 &  83.92 \\
     (j) & \CheckmarkBold &   & \CheckmarkBold & \CheckmarkBold &   &          63.15$\pm$0.45 &    72.95 &    47.48 &  62.84 &  94.96 &  83.78 \\
     (k) & \CheckmarkBold &   & \CheckmarkBold &   & \CheckmarkBold & \textbf{64.59}$\pm$0.17 & \textbf{74.03} & \textbf{49.60} & \textbf{64.12} & 95.00 &  \textbf{84.27} \\
    \bottomrule
    \end{tabular}
    \caption{
        \fscore scores on the NTC 1.5 development set on \textbf{\textit{inter-sentential}} setting. 
        Bold value indicates the best results in the same column. PT and FT are abbreviations of pretraining and finetuning.
    }
    \label{tab:inter-dev-score}
\end{table*}
%%%%%

%%%%%%%%%%
%%%%%%%%%%
\section{Number of Parameters of each Model}
We report total number of parameters of \prev{} and \proposed{} in Table~\ref{tab:appendix-model-num-param}.
%%%%%
\begin{table}[t]
    \centering
        \footnotesize
        \begin{tabular}{l|r}
        \toprule
        Model & Number of Parameters\\
        \midrule
        \prev{}             &  111,223,312 \\
        \proposed{}         &  112,395,268 \\
        \bottomrule
        \end{tabular}
    \caption{Number of parameters of the models}
    \label{tab:appendix-model-num-param}
\end{table}
%%%%%
\begin{table*}[t]
    \centering
    \footnotesize
    \begin{tabular}{l|cccc|c|c}
    \toprule
            & \multicolumn{4}{c|}{ZAR} & DEP & All \\
    Method  & All & \textit{intra} & \textit{inter} & \textit{exophoric} & & \\
    \midrule
    \multicolumn{7}{c}{\textbf{Finetuning}} \\
    \midrule
    \prev{} &   62.27$\pm$0.42 &       71.55 &          44.30 &           64.04 &           94.44 &           82.97 \\
    \proposed{} &  62.47$\pm$0.53 &         71.09 &          45.20 &           64.41 &           94.46 &           83.03 \\
    \midrule
    \multicolumn{7}{c}{\textbf{10K Further Pretraining + Finetuning}} \\
    \midrule    
    \textit{Cloze} + \prev{} & 62.52$\pm$0.37 &         71.40 &          44.79 &           64.55 &           94.48 &           83.08 \\
    \textit{Cloze} + \proposed{} & 62.42$\pm$0.30 &         71.04 &          45.31 &           64.15 &           94.49 &           83.04 \\
    \task{} + \prev{} & 62.63$\pm$0.41 &         71.63 &          45.14 &           64.47 &           94.39 &           83.07 \\
    \task{} + \proposed{} & \textbf{63.52}$\pm$0.19 & \textbf{72.07} & \textbf{47.43} &  \textbf{64.95} &   \textbf{94.58} &  \textbf{83.46} \\
    \midrule
    \multicolumn{7}{c}{\textbf{30K Further Pretraining + Finetuning}} \\
    \midrule    
    \textit{Cloze} + \prev{} & 62.54$\pm$0.47 &         71.82 &          44.98 &           63.94 &  \textbf{94.51} &           83.10 \\
    \textit{Cloze} + \proposed{} & 62.85$\pm$0.19 &         71.52 &          45.97 &           64.55 &           94.49 &           83.18 \\
    \task{} + \prev{} & 63.06$\pm$0.19 &         71.96 &          46.37 &           64.42 &           94.43 &           83.26 \\
    \task{} + \proposed{} & \textbf{64.18}$\pm$0.23 & \textbf{72.67} & \textbf{48.41} & \textbf{65.40} &  94.50 &  \textbf{83.65} \\
    \bottomrule
    \end{tabular}
    \caption{\fscore scores on the NTC 1.5 \textbf{test set}. Bold value indicates the best results in the same column.}
    \label{tab:appendix-test}
\end{table*}
%%%%%

\end{document}